\pdfoutput=1

\documentclass[11pt]{article}

\usepackage[dvipsnames]{xcolor}
\usepackage{emnlp2021}

\usepackage{times}
\usepackage{latexsym}

\usepackage[T1]{fontenc}

\usepackage[utf8]{inputenc}

\usepackage{microtype}

\usepackage{microtype}
\usepackage{tablefootnote}
\usepackage{graphicx}
\graphicspath{ {./images/} }
\usepackage{caption}
\usepackage{multirow}
\usepackage{multicol}
\usepackage{array}
\usepackage{subcaption}

\newcommand{\ie}{\textit{i.e., }}
\newcommand{\eg}{\textit{e.g., }}

\title{TIAGE: A Benchmark for Topic-Shift Aware Dialog Modeling}

\author{Huiyuan Xie$^1$ \quad Zhenghao Liu$^2$ \quad Chenyan Xiong$^3$ \quad Zhiyuan Liu$^4$ \quad Ann Copestake$^1$ \\
$^1$Department of Computer Science and Technology, University of Cambridge, UK \\
$^2$Department of Computer Science and Technology, Northeastern University, China \\
$^3$Microsoft Research, United States \\
$^4$Department of Computer Science and Technology, Tsinghua University, China\\
\texttt{\{hx255,aac10\}@cl.cam.ac.uk, liuzhenghao@cse.neu.edu.cn}\\
\texttt{chenyan.xiong@microsoft.com, liuzy@tsinghua.edu.cn}
}

\begin{document}

\maketitle

\begin{abstract}

Human conversations naturally evolve around different topics and fluently move between them. In research on dialog systems, the ability to actively and smoothly transition to new topics is often ignored. In this paper we introduce \textsc{TIAGE}, a new topic-shift aware dialog benchmark constructed utilizing human annotations on topic shifts. Based on \textsc{TIAGE}, we introduce three tasks to investigate different scenarios of topic-shift modeling in dialog settings: topic-shift detection, topic-shift triggered response generation and topic-aware dialog generation. Experiments on these tasks show that the topic-shift signals in \textsc{TIAGE} are useful for topic-shift response generation. On the other hand, dialog systems still struggle to decide when to change topic. This indicates further research is needed in topic-shift aware dialog modeling.\footnote{Code and data available at: https://github.com/HuiyuanX\\ie/tiage.}

\end{abstract}

\section{Introduction}
Existing dialog models \cite{ghandeharioun2019approximating,einolghozati2019improving,liu2018dialogue} have been reported to perform well in generating on-topic utterances in dialog scenarios. However, those models still struggle to proactively generate appropriate topic-shift utterances in conversations~\cite{holtzman2019curious, zhang2020grounded}. 

\begin{figure}[t]
\centering
\includegraphics[width=0.8\linewidth]{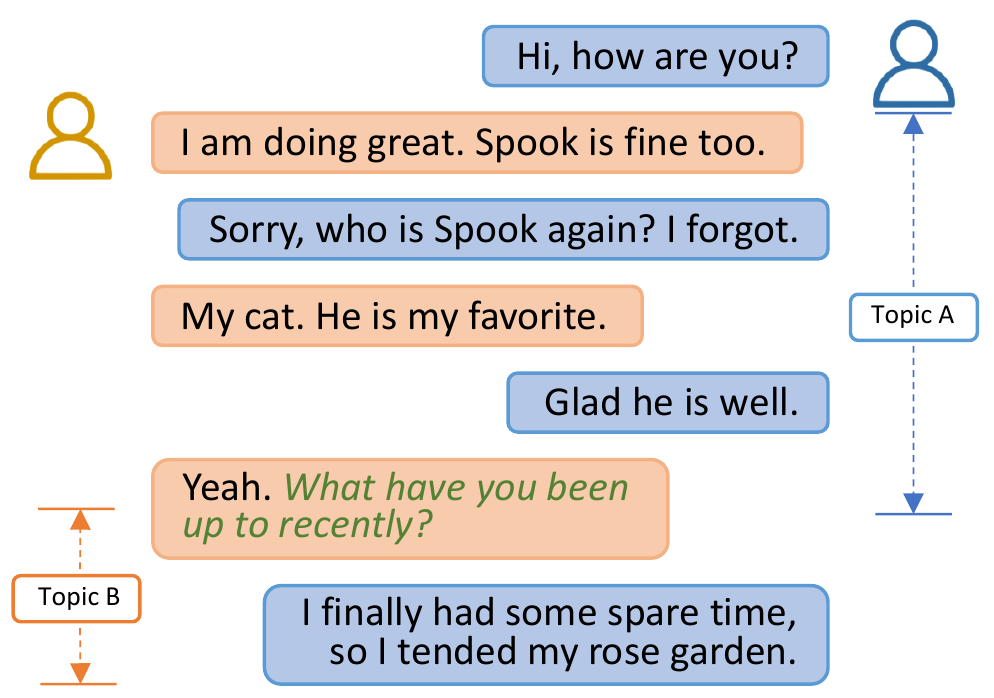}
\caption{An example of topic-shift behaviors in human conversations. Topic-shift utterances are highlighted in \textcolor{ForestGreen}{green} and in \textit{italic}. Changing the topic helps keep the conversation going on.}
\label{fig:ts_dialog}
\end{figure}

It is beneficial for dialog systems to be able to shift topics fluently. As shown in Figure \ref{fig:ts_dialog}, topic-shift behaviors are commonly observed in human conversations~\cite{brown1983discourse}. Fluent topic shifts therefore are crucial for dialog models to be able to model or mimic human conversational patterns. Proactively using topic shifts can help chatbots guide conversations to a pre-defined target~\cite{tang2019target}. Furthermore, switching topics allows chatbots to maintain engaging conversations with users. Without the ability to actively shift topics away from tired topics, chatbots risk generating dull responses or repeating themselves regarding a specific topic.

To facilitate research on topic-shift dialog modeling, we curate a Topic-shIft Aware dialoG datasEt (\textsc{TIAGE}) by augmenting the PersonaChat dataset \cite{zhang2018personalizing} with topic-shift annotations. To the best of our knowledge, \textsc{TIAGE} is the first dataset that focuses on topic-shift behaviors in open-domain dialog data. 
\textsc{TIAGE} contains a human annotated dataset with 7,861 gold standard topic-shift annotations, and a weak supervision dataset to adapt pretrained NLG systems to PersonaChat-style data. The inter-annotator agreement for topic-shift annotations in \textsc{TIAGE} is 0.479.

With \textsc{TIAGE}, we propose three tasks to study topic-shift behaviors: topic-shift detection, topic-shift triggered response generation and topic-aware dialog generation. The topic-shift detection task asks models to detect whether the ongoing topic has shifted or should shift. The other two tasks focus on modeling topic-shift behaviors in response generation. Specifically, the topic-shift triggered response generator receives a fixed topic-shift signal to generate topic-shift responses, whilst the topic-aware dialog generation task requires dialog systems to predict the topic-shift trigger by themselves.

Our experiments reveal that the topic-shift signals in \textsc{TIAGE} indeed improve dialog systems' ability to generate topic-shift responses. However, it is difficult for dialog models to predict when it is appropriate to change topics. These observations highlight the need for better modeling of topic shifts in dialog generation. We hope our benchmark can motivate further research on topic-shift aware dialog modeling.

\section{Related Work}


Existing work in dialog systems falls into two broad categories. \textit{Task-oriented} dialog systems \cite{budzianowski2018multiwoz,liu2018dialogue} help users complete tasks in specific domains. \textit{Open-domain} dialog systems \cite{chen2017open,tang2019target} allow agents to have open-ended conversations with users. Most existing dialog models \cite{fang2018sounding,zhang2019dialogpt,ghandeharioun2019approximating} emphasize end-to-end response generation, and do not explicitly address the topic-shift problem in dialog generation. 

Early work in topic detection and segmentation \cite{hirschberg1993empirical,passonneau1997discourse} focused on identifying cue phrases (such as \textit{on a different note}) or examining lexical cohesion to segment topical chunks. Other work \cite{fiscus2002topic} investigated topic detection and tracking (TDT) in a stream of broadcast news stories. More recent work \cite{somasundaran2020two} has explored utilizing neural networks to address topic segmentation. Although some of the existing work \cite{galley2003discourse,arnold2019sector} has investigated topic detection in dialog-style data, the generation aspect of topic-shift modeling in dialog settings is still unclear.

\section{Topic-Shift Aware Dialog Dataset}
\label{sec:dataset}

In this section we introduce the rationale for our choice of data source, the human annotation process of topic-shift labelling in \textsc{TIAGE} and its data statistics. We also analyze the linguistic patterns of topic-shift utterances in \textsc{TIAGE}.

\textbf{Rationale for our choice of data source.}
We construct \textsc{TIAGE} by augmenting the PersonaChat dataset \cite{zhang2018personalizing} with topic-shift human annotations. We view PersonaChat as a suitable dataset for topic-shift annotation for the following reasons: (1) the Personachat data was collected online in a textual form by mimicking chit-chat scenarios, where natural shifts of topics are more likely to happen; (2) dialogs in this dataset contain more than 10 dialog turns, and longer dialog contexts tend to exhibit a conversational flow with more topics; and (3) despite the fact that some participants in PersonaChat may have rushed into changing topics to quickly exchange their profile information, we observed that most of the participants still manage to change topics in a more natural and coherent way, making this dataset a favorable choice to study topic-shift behaviors.

\begin{table}[t]
\small
\centering
\begin{subtable}[h]{0.5\textwidth}
    \centering
    \begin{tabular}{lll}
    \hline
        & \textbf{\textsc{WeakSupo}$_{train}$} & \textbf{\textsc{WeakSupo}$_{dev}$} \\ \hline
        \#Dialogs & 7,939 & 1,000 \\
        \#Instances & 108,711 & 13,788 \\ 
        \#AvgTurns & 14.7 & 14.8 \\\hline
    \end{tabular}
    \caption{The weak supervision data split.}
    \label{tab:ws_data}
\end{subtable}
\\[0.3cm]
\begin{subtable}[h]{0.5\textwidth}
    \centering
    \begin{tabular}{llll}
    \hline
        & \textbf{\textsc{Anno}$_{train}$} & \textbf{\textsc{Anno}$_{dev}$} & \textbf{\textsc{Anno}$_{test}$} \\ \hline
        \#Dialogs & 300 & 100 & 100 \\
        \#Instances & 4,767 & 1,546 & 1,548 \\ 
        \#AvgTurns & 15.6 & 15.5 & 15.6 \\ \hline
    \end{tabular}
    \caption{The human annotated data split.} 
    \label{tab:anno_data}
\end{subtable}

\caption{Data statistics. \textit{\#AvgTurns} denotes the average number of turns per dialog. Each \textit{instance} is a (context, response) pair around a specific dialog turn. The average number of tokens per utterance is 11.8. In the human annotated data split, the average number of topic-shift turns per dialog is 3.5. The vocabulary size of the entire dataset is around 18K.}
\label{tab:data_stats}
\end{table}

\textbf{Human annotation process.}
For the annotation pool, we have a total number of 25 human annotators. We randomly selected 500 dialogs from the original PersonaChat dev/test datasets, resulting in 7,861 dialog turns to label. Each dialog turn was randomly assigned to and independently labeled by 2 annotators. For each dialog turn, we asked annotators to indicate whether they think the conversational topic is changed at that turn. During the annotation process, all annotators were talked through the general aim of this annotation task and given the same annotation guidelines (see Appendix \ref{app:human_anno} for details). 

Since topic is co-constructed, it is rather limiting to analyze a turn for itself when trying to identify topic transitions. To facilitate the recognition of slowly transitioned topics, we encouraged the annotators to take into account both the previous two turns and the following two turns of the target dialog turn to make a decision. This helped decision making for cases where topics are slowly developed and transitioned.

After annotating, we obtained a dialog dataset with gold standard topic-shift labels for 7,861 dialog turns. The Cohen's Kappa score for all annotations is 0.479\footnote{The Cohen's Kappa score ranges from 0.41 to 0.60 indicating moderate agreement, which confirms the quality of the human annotations of \textsc{TIAGE}.}. Annotated examples of \textsc{TIAGE} dialogs are shown in Appendix \ref{app:tiage_examples}.

\textbf{Dataset statistics.} 
As shown in Table \ref{tab:data_stats}, \textsc{TIAGE} provides \textit{weak supervision data} and \textit{human annotated data} to train dialog models. \textit{Weak supervision data} is selected from the original PersonaChat training set and helps adapt NLG models to PersonaChat-style data. The weak supervision data consists of 8,939 dialogs and is split into two sets: \textsc{WeakSupo}$_{train}$ and \textsc{WeakSupo}$_{dev}$. \textit{Human annotated data} consists of 500 annotated dialogs with topic-shift annotations at each dialog turn. We split them into 300 \textsc{Anno}$_{train}$, 100 \textsc{Anno}$_{dev}$ and 100 \textsc{Anno}$_{test}$ dialogs respectively. As each dialog has multiple dialog turns, we extract (context, response) pairs as instances for all turns in each dialog.

\textbf{Analysis of topic-shift patterns.}
We examine a number of topic-shift utterances labeled by human annotators. We find that many of the topic-shift responses demonstrate an interesting pattern of [\textit{comment}; \textit{topic shift}]. More specifically, the response that changes the conversational topic is typically a brief \textit{comment} on the previous dialog context, tailed by a \textit{topic-shift} sentence with a different conversational focus. The \textit{comment} usually corresponds to the sentiment previously expressed in the dialog.

This pattern echoes some of the findings in pragmatics research~\cite{brown1987politeness,goldsmith2007brown}. When speakers introduce a new topic, it is a common positive politeness strategy \cite{leech2014pragmatics} to first respond to the content uttered by other speakers. This pattern is potentially useful for dialog systems seeking to generate topic-shift utterances in a natural and coherent manner. Before introducing a new topic, it is favorable for dialog systems to first generate a comment regarding the previous topic that expresses either \textit{approbation} (\eg ``great'', ``that's cool'') or \textit{sympathy} (\eg ``that's too bad'' or ``I'm sorry to hear that''). This shows that they are attuned to the users' interests and needs.

\section{Tasks of Topic-Shift Modeling}

Along with dialog utterances, \textsc{TIAGE} also provides gold standard topic-shift labels for dialog turns. This enables us to model topic shifts in dialog scenarios. We first introduce two tasks: topic-shift detection and topic-shift triggered response generation. They can be considered as intermediate steps of the topic-aware dialog generation task.

\subsection{Preliminary of Response Generation}

When considering a specific turn in a dialog, we denote the current utterance and all its previous utterances as the \textit{context} $\mathbf{X}_T=\{\mathbf{x}_1, ..., \mathbf{x}_i, ..., \mathbf{x}_N\}$ where $\mathbf{x}_i$ is the $i$-th utterance in the dialog history, and $N$ is the context length. Then we expect the \textit{response} to be generated after the current utterance $\mathbf{x}_N$. We denote a \textit{topic-shift response} as $\mathbf{s}_{TS}=\{s_1, ..., s_i, ..., s_T\}$ where $s_i$ is the $i$-th token in the response and $T$ is the sentence length. Similarly, an \textit{on-topic response} is denoted as $\bar{\mathbf{s}}_{NTS}=\{\bar{s}_1, ..., \bar{s}_i, ..., \bar{s}_M\}$ where $\bar{s}_i$ is the $i$-th token in the response and $M$ is the sentence length.

\subsection{Topic-Shift Detection}
Topic-shift detection is a fundamental task that evaluates models' ability to detect topic-shift occurrence at dialog turns. 

\textbf{Task definition.} We introduce two settings for this task. In the \textit{retrospective} setting, models have access to both the dialog context $\mathbf{X}_T$ and the corresponding response (either $\mathbf{s}_{TS}$ or $\bar{\mathbf{s}}_{NTS}$) to detect topic-shift occurrence, whilst in the \textit{predictive} setting, models are asked to make topic-shift predictions based on the context $\mathbf{X}_T$ only.

\textbf{Topic shift classifiers.}
We first implement three retrospective classifiers. We employ \texttt{GenEnc} which uses the GEN Encoder \cite{zhang2019generic} to separately encode dialog context and response into embeddings to estimate the topic-shift intents. \texttt{GenEnc} uses a cosine similarity threshold of 0.25 to filter out (context, response) pairs, and classify them as topic-shift occurrences. Then we implement a \texttt{BERT-Wiki727k} model \cite{devlin2018bert} trained on the \textsc{Wiki-727k} dataset \cite{koshorek2018text}. We also employ a T5 model~\cite{raffel2019exploring} finetuned on the \textsc{Anno}$_{train}$ data with topic-shift labels as our retrospective T5 topic-shift classifier (denoted as \texttt{RetroTS-T5}).

For the predictive setting, we implement a T5-based topic-shift manager (denoted as \texttt{TSManager}) and finetune it on the \textsc{Anno}$_{train}$ data. The major difference between \texttt{RetroTS-T5} and \texttt{TSManager} is that \texttt{RetroTS-T5} has access to both the dialog context and the response, while \texttt{TSManager} makes topic-shift predictions based solely on the context.

\subsection{Topic-Shift Triggered Response Generation}
This task examines models' ability to generate topic-shift utterances when a need to change topics is signaled. 

\textbf{Task definition.}
Given a dialog context $\mathbf{X}_T$, the topic-shift triggered response generation task requests models to directly generate a response $\mathbf{s}_{TS}$ that shifts the conversation to a different topic.

\textbf{Topic-shift triggered generator.}
We build a \texttt{T5-NLG}$_{TS}$ response generator using the pretrained T5 model. We first train the T5 model on the \textsc{WeakSupo}$_{train}$ data, and then further finetune it on the topic-shift instances (\ie where topic shifts occur) in the \textsc{Anno}$_{train}$ data. 

\textbf{Compared approaches.}
We also try a number of topic-insensitive NLG models for comparison. We train a \texttt{T5-NLG} model on the \textsc{WeakSupo}$_{train}$ data without any topic-shift signals. We use the \texttt{DialoGPT} model~\cite{zhang2019dialogpt} finetuned on the same data as another baseline.

\subsection{Topic-Aware Dialog Generation}

The third task we propose targets more difficult and realistic topic-shift modeling in dialog generation.

\textbf{Task definition.}
More formally, given a dialog context $\mathbf{X}_T$, the goal of the topic-aware dialog generation task is to generate a topic-shift response $\mathbf{s}_{TS}$ if a change of topic is needed, or an on-topic response $\bar{\mathbf{s}}_{NTS}$ if otherwise. The topic-aware dialog generation task asks models to identify the need to change topics by themselves and generate topic-shift or on-topic responses according to the prediction. 

\textbf{Topic-aware dialog system.}
Our topic-aware dialog system (\texttt{TADial}) is a pipeline system. We separately train two T5-based response generators: \texttt{T5-NLG}$_{TS}$ and \texttt{T5-NLG}$_{NTS}$. We switch between the two response generators to produce either a topic-shift or on-topic response, guided by the topic-shift signals from \texttt{TSManager}. \texttt{T5-NLG}$_{TS}$ aims to generate topic-shift responses, while \texttt{T5-NLG}$_{NTS}$ is finetuned on non-topic-shift instances to generate on-topic responses.

\textbf{Compared approaches.}
We use the \texttt{T5-NLG} and \texttt{DialoGPT} models finetuned on the \textsc{WeakSupo}$_{train}$ data as baselines for comparison.

\section{Evaluation Results}
We report here the evaluation results for baseline systems on the above three tasks.

\begin{table}[t]
\small
    \centering
    \begin{tabular}{c|ccc}
    \hline
        \textbf{Approaches} & \textbf{Precision} & \textbf{Recall} & \textbf{F1-score} \\ \hline
        \textsc{BERT-Wiki727k} & 0.412 & 0.020 & 0.038 \\ 
        \textsc{GenEnc} & 0.337 & 0.199 & 0.250 \\ 
        \textsc{RetroTS-T5} & 0.709 & 0.657 & 0.682 \\ \hline
        \textsc{TSManager} & 0.340 & 0.170 & 0.220 \\ \hline\hline
        \textbf{Human\footnotemark} & 0.687 & 0.607 & 0.644 \\ \hline
    \end{tabular}
    \caption{Model performance on the topic-shift detection task.}
    \label{tab:p_r_f1}
\end{table}
\footnotetext[2]{We use the annotations from one annotator as gold standard references, and calculate human performance on the annotations from the other annotator.}

\textbf{Topic-shift detection.}
We test topic-shift classifiers on the annotated \textsc{Anno}$_{test}$ split. From Table \ref{tab:p_r_f1} we observe that \texttt{RetroTS-T5} outperforms other approaches by a large margin and is on par with human performance. This indicates that topic shifts in PersonaChat dialogs exhibit certain patterns, which can be captured from our human-labeled topic-shift annotations by our retrospective T5 classifier. We also notice that there is a clear gap in classification performance between \texttt{RetroTS-T5} and \texttt{TSManager}. The predictive setting of \texttt{TSManager} is inherently harder than \texttt{RetroTS-T5}, as it is asked to predict topic-shift labels based solely on dialog context. 

\begin{table}[t]
\centering
\resizebox{0.48\textwidth}{!}{
\begin{tabular}{c|cccc}
\hline 
\textbf{Model} & \textbf{BLEU-2} & \textbf{METEOR} & \textbf{ROUGE\_L} & \textbf{CIDEr}\\ 
\hline
\textsc{DialoGPT} & 0.060 & 0.077 & 0.125 & 0.104 \\
\textsc{T5-NLG} & 0.079 & 0.086 & 0.161 & 0.170 \\
\textsc{T5-NLG}$_{TS}$ & 0.092 & 0.092 & 0.177 & 0.175 \\
\hline
\end{tabular}}
\caption{Evaluation results of topic-shift triggered response generation on topic-shift instances in \textsc{Anno}$_{test}$.}
\label{tab:ts_eval}
\end{table}

\textbf{Topic-shift triggered response generation.}
In Table \ref{tab:ts_eval} we report evaluation results\footnote{We use the \texttt{nlg-eval} package for automatic evaluation. https://github.com/Maluuba/nlg-eval.} of our topic-shift triggered response generator (\texttt{T5-NLG}$_{TS}$) and two topic-insensitive models (\texttt{DialoGPT} and \texttt{T5-NLG}). Models are tested on the \textit{topic-shift} instances in \textsc{Anno}$_{test}$. We observe that \texttt{T5-NLG} yields better performance than \texttt{DialoGPT}. Furthermore, \texttt{T5-NLG}$_{TS}$ achieves better evaluation results on topic-shift test instances, outperforming \texttt{T5-NLG} by 16.46\% in BLEU-2 and 9.94\% in ROUGE\_L. 
The better performance of \texttt{T5-NLG}$_{TS}$ validates the effectiveness of topic-shift signals in improving topic-shift response generation. It also proves that explicitly modeling topic-shift behaviors can potentially benefit dialog generation.

\begin{table}[t]
\centering
\resizebox{0.48\textwidth}{!}{
\begin{tabular}{c|cccc}
\hline 
\textbf{Model} & \textbf{BLEU-2} & \textbf{METEOR} & \textbf{ROUGE\_L} & \textbf{CIDEr}\\ 
\hline
\textsc{DialoGPT} & 0.063 & 0.077 & 0.134 & 0.125 \\
\textsc{T5-NLG} & 0.082 & 0.087 & 0.159 & 0.175 \\
\textsc{TADial} & 0.082 & 0.087 & 0.162 & 0.177 \\
\hline
\end{tabular}}
\caption{Evaluation results of topic-aware dialog generation on all instances in \textsc{Anno}$_{test}$.}
\label{tab:all_eval}
\end{table}

\textbf{Topic-aware dialog generation.}
We test \texttt{TADial} and two topic-insensitive baselines on \textit{all} instances in \textsc{Anno}$_{test}$. From Table \ref{tab:all_eval}, we can see that \texttt{TADial} with a dedicated topic-shift management component does not yield better performance over the \texttt{T5-NLG} model which is simply trained on dialog instances with no topic-shift labels. This points out that due to the deficiency of \texttt{TSManager} signals, hard-wiring a topic-shift management component into the generation pipeline falls short to improve generation results. It remains a challenging task to produce well-timed and good-quality topic-shift signals based on dialog context only, which hinders overall topic-aware dialog generation.

\section{Conclusion and Future Work}

We construct the \textsc{TIAGE} dataset with human annotated topic-shift labels on the basis of the PersonaChat dataset. Based on \textsc{TIAGE}, we introduce three tasks: topic-shift detection, topic-shift triggered response generation and topic-aware dialog generation. Empirical results show that topic-shift labels in \textsc{TIAGE} are useful for topic-shift response generation. 
However, it remains a challenging task for dialog models to predict good-quality topic-shift signals based on dialog context only. Further research is needed on selecting appropriate topics to shift to among multiple references. Natural topic shifts can be both a precaution against, and a remedy to, dull and repetitive response generation in real-world dialog applications. \textsc{TIAGE} with its topic-shift annotations can help direct future investigation on the incorporation of topic-shift tactics in dialog models, which allows more effective control over topic-shift aware dialog generation. 

\section*{Acknowledgments}
We thank the anonymous reviewers for their constructive feedback. We thank our annotators from the University of Cambridge and Tsinghua University for their help in annotating the topic-shift labels used in this paper. Huiyuan Xie is grateful for being supported by the CSC Cambridge Scholarship. Zhenghao Liu is supported by National Natural Science Foundation of China (NSFC) under grant No. 61872074 and 61772122. This work is partly supported by the National Key Research and Development Program of China (No. 2020AAA0106501).

\bibliography{anthology,custom}
\bibliographystyle{acl_natbib}

\clearpage
\appendix
\section{Appendix}

\subsection{Human Annotation Guidelines}
\label{app:human_anno}

Here we present the annotation guidelines used for the human annotation process in this work.

\textbf{Task description.}
Chitchat systems are expected to have the ability to proactively change conversational topics when necessary. For occasions when a chat agent runs out of things to say or the current discussion is starting to get boring, topic shifting is a common tactic to keep the conversation going on. In this work, we aim to model topic-shift phenomenon in open-domain dialog settings. To achieve this, we need to construct a new dialog dataset with topic-shift signals.

\begin{table}[ht]
\small
\begin{subtable}[h]{0.5\textwidth}
    \centering
    \begin{tabular}{lp{5cm}}
    \hline
    [Speaker1:] & My dad works for the New York Times. \\
    
    [Speaker2:] & Oh wow! You know, I dabble in photography; maybe you can introduce us sometime. \\
    
    [Speaker1:] & \textbf{\textit{Photography is the greatest art out there.}} \textcolor{NavyBlue}{$\rightarrow$ not a topic shift} \\
    \hline
    \end{tabular}
    \label{tab:examples_1}
    \caption{Commenting on the previous context.}
\end{subtable}
\\[0.3cm]
\begin{subtable}[h]{0.5\textwidth}
    \centering
    \begin{tabular}{lp{5cm}}
    \hline
    [Speaker1:] & Do you teach cooking?  \\
    
    [Speaker2:] & \textbf{\textit{No, since I'm a native of Mexico, I teach Spanish.}} \textcolor{NavyBlue}{$\rightarrow$ not a topic shift} \\
    \hline
    \end{tabular}
    \label{tab:examples_2}
    \caption{Question answering.}
\end{subtable}
\\[0.3cm]
\begin{subtable}[h]{0.5\textwidth}
\centering
    \begin{tabular}{lp{5cm}}
    \hline
    
    [Speaker1:] & Pets are cute! \\
    
    [Speaker2:] & \textbf{\textit{I heard that Huskies are difficult dogs to take care of.}} \textcolor{NavyBlue}{$\rightarrow$ not a topic shift} \\
    \hline
    \end{tabular}
    \label{tab:examples_3a}
    \caption{Developing the conversation to sub-topics.}
\end{subtable}
\\[0.3cm]
\begin{subtable}[h]{0.5\textwidth}
\centering
    \begin{tabular}{lp{5cm}}
    \hline
    [Speaker1:] & You are an artist? What kind of art, I do American Indian stuff. \\
    
    [Speaker2:] & Yes, I love to draw. \textbf{\textit{I love to eat too, sometimes too much.}} \textcolor{NavyBlue}{$\rightarrow$ topic shift} \\
    \hline
    \end{tabular}
    \caption{Introducing a relevant but different topic.}
    \label{tab:examples_3b}
\end{subtable}
\\[0.3cm]
\begin{subtable}[h]{0.5\textwidth}
\centering
    \begin{tabular}{lp{5cm}}
    \hline
    [Speaker1:] & What do you do for fun? \\
    
    [Speaker2:] & I drive trucks so me and my buds go truckin in the mud. \\
    
    [Speaker1:] & Must be fun! My version of that's running around a library! \\
    
    [Speaker2:] & \textbf{\textit{Do you have a favourite animal? Chickens are my favourite. I love them.}} \textcolor{NavyBlue}{$\rightarrow$ topic shift} \\
    \hline
    \end{tabular}
    \caption{Completely changing the topic.}
    \label{tab:examples_4}
\end{subtable}

\caption{Different scenarios of dialog response in conversations.}
\label{tab:gen_examples}
\end{table}

\textbf{Data annotation.}
For each utterance in a dialog, annotators are asked to decide whether the topic of the conversation changes when transiting from the current utterance to the following response. If there is a topic shift, annotators should label the response with ``1'', otherwise label it with ``0''.

In conversations, the response of a speaker to the dialog context usually falls into one of the following cases (see examples in Table \ref{tab:gen_examples}):

(a) Commenting on what the other participant just said (the most common scenario);

(b) Question answering;

(c) Developing the conversation to sub-topics;

(d) Introducing a relevant but different topic;

(e) Completely changing the topic.

\textbf{Other tips for data labeling.}
A number of words and phrases are often used as indicators for topic shifts, including but not limited to: but, speaking of, talking about, anyway, by the way, that reminds me, before I forget, I want to mention, let's talk about, we need to discuss, funny you should mention that, not to change the subject but, changing the topic slightly, totally unrelated, on a different/relevant note.

\subsection{Examples of Labeled Data in \textsc{TIAGE}}
\label{app:tiage_examples}

In Table \ref{tab:labeled_data} we showcase examples of labeled dialogs selected from \textsc{TIAGE}.
 
\begin{table}[t]
\small
\begin{subtable}[h]{0.5\textwidth}
    \centering
    \begin{tabular}{lp{4cm}|c}
    \hline
    & \textbf{Dialog} & \textbf{TS Label} \\ \hline
    [Speaker1:] & Hi! How are you this evening? & N/A \\ \hline
    
    [Speaker2:] & Good. I spent all afternoon walking my dogs. I've three Labradors. & \textbf{0} \\ \hline
    
    [Speaker1:] & Cool, that's a lot of dogs. Do you like music? I love it. & \textbf{1} \\
    \hline
    \end{tabular}
\end{subtable}
\\[0.3cm]
\begin{subtable}[h]{0.5\textwidth}
    \centering
    \begin{tabular}{lp{4cm}|c}
    \hline
    & \textbf{Dialog} & \textbf{TS Label} \\ \hline
    [Speaker1:] & I think you are great. You are my best friend. & N/A \\ \hline
    
    [Speaker2:] & My best friend is a bear, bears don't have friends, that's why they're my favourite. & \textbf{0} \\ \hline
    
    [Speaker1:] & Webster's dictionary defines weddings as the fusing of two metals with a hot torch. & \textbf{1} \\
    \hline
    \end{tabular}
\end{subtable}
\caption{Annotated dialog examples in \textsc{TIAGE}.}
\label{tab:labeled_data}
\end{table}


\subsection{Implementation Details}
For the topic-shift classifiers, we use the base version of BERT and T5 models, initialized from their pretrained weights. For the dialog response generation experiments, we use the small version of DialoGPT and the base version of T5. Our implementation is based on the HuggingFace Transformers library \cite{wolf2020transformers}. All models are optimized using Adam with a learning rate of 5e-5 and a batch size of 64. We set the maximum input sequence length to 512. The training is carried out using 1 Nvidia RTX 8000 GPU and takes around 15 hours.


    
    
    
    
    
    


\end{document}